\documentclass[letterpaper, 10 pt, journal, twoside]{IEEEtran}
\IEEEoverridecommandlockouts                             

\usepackage{amssymb}

\usepackage{subfig}
\usepackage{multicol}
\usepackage[bookmarks=true]{hyperref}
\usepackage{graphicx,color,subfig}
\usepackage{amsfonts, amssymb, amsmath,verbatim, multirow}
\usepackage{enumerate}
\usepackage{breqn}
\usepackage{amsmath}
\usepackage{diagbox}
\usepackage{times}
\usepackage{algorithm}
\usepackage{algorithmic}
\usepackage{easybmat}
\usepackage{balance}

\usepackage{caption}

\newcommand{\rulesep}{\unskip\ \vrule\ }

\newcommand{\mb}[1]{{\boldsymbol{#1}}}

\renewcommand{\UrlFont}{\small}
\newcommand{\todo}[1]{\textcolor{blue}{#1}}

\usepackage[usenames,dvipsnames]{xcolor}

\usepackage{verbatim}  

\title{A Fog Robotics Approach to Deep Robot Learning: Application to Object Recognition and Grasp Planning in Surface Decluttering}
\author{Ajay Kumar Tanwani, Nitesh Mor, John Kubiatowicz, Joseph E. Gonzalez, Ken Goldberg%
\thanks{The AUTOLAB at UC Berkeley (\UrlFont{automation.berkeley.edu}).
}
\thanks{University of California, Berkeley.
        {\tt \{ajay.tanwani, mor, kubitron, jegonzal, goldberg\}@berkeley.edu}}%
}

\begin{document}

\maketitle


\begin{abstract}
The growing demand of industrial, automotive and service robots presents a challenge to the centralized Cloud Robotics model in terms of privacy, security, latency, bandwidth, and reliability. In this paper, we present a `Fog Robotics' approach to deep robot learning that distributes compute, storage and networking resources between the Cloud and the Edge in a federated manner. Deep models are trained on non-private (public) synthetic images in the Cloud; the models are adapted to the private real images of the environment at the Edge within a trusted network and subsequently, deployed as a service for low-latency and secure inference/prediction for other robots in the network. We apply this approach to surface decluttering, where a mobile robot picks and sorts objects from a cluttered floor by learning a deep object recognition and a grasp planning model. Experiments suggest that Fog Robotics can improve performance by sim-to-real domain adaptation in comparison to exclusively using Cloud or Edge resources, while reducing the inference cycle time by $4\times$ to successfully declutter $86\%$ of objects over $213$ attempts.
\end{abstract}


\section{INTRODUCTION}

The term `Cloud Robotics' describes robots or automation systems that rely on either data or code from the Cloud, i.e. where not all sensing, computation, and memory is integrated into a single standalone system \cite{Kuffner10,Kehoe15}. By moving the computational and storage resources to the remote datacenters, Cloud Robotics facilitates sharing of data across applications and users, while reducing the size and the cost of the onboard hardware. Examples of Cloud Robotics platforms include RoboEarth \cite{Waibel11}, KnowRob \cite{Tenorth13}, RoboBrain \cite{Saxena14}, DexNet as a Service \cite{Tian17,Li_CASE_18}. Recently, Amazon RoboMaker~\cite{RoboMaker18} and Google Cloud Robotics~\cite{GoogleCR_18} released platforms to develop robotic applications in simulation with their Cloud services.


Robots are increasingly linked to the network and thus not limited by onboard resources for computation, memory, or software. Internet of Things (IoT) applications and the volume of sensory data continues to increase, leading to a higher latency, variable timing, limited bandwidth access than deemed feasible for modern robotics applications~\cite{Goldberg19,Wan16}. Moreover, stability issues arise in handling environmental uncertainty with any loss in network connectivity. Another important factor is the security of the data sent and received from heterogeneous sources over the Internet. The correctness and reliability of information has direct impact on the performance of robots. Robots often collect sensitive information (e.g., images of home, proprietary warehouse and manufacturing data) that needs to be protected. As an example, a number of sensors and actuators using Robot Operating System (ROS) have been exposed to public access and control over the Internet~\cite{DeMarinis18}. 

\begin{figure}[tbp]
\centering
\includegraphics[trim={0cm 0cm 0cm 0cm},clip,scale = 0.68]{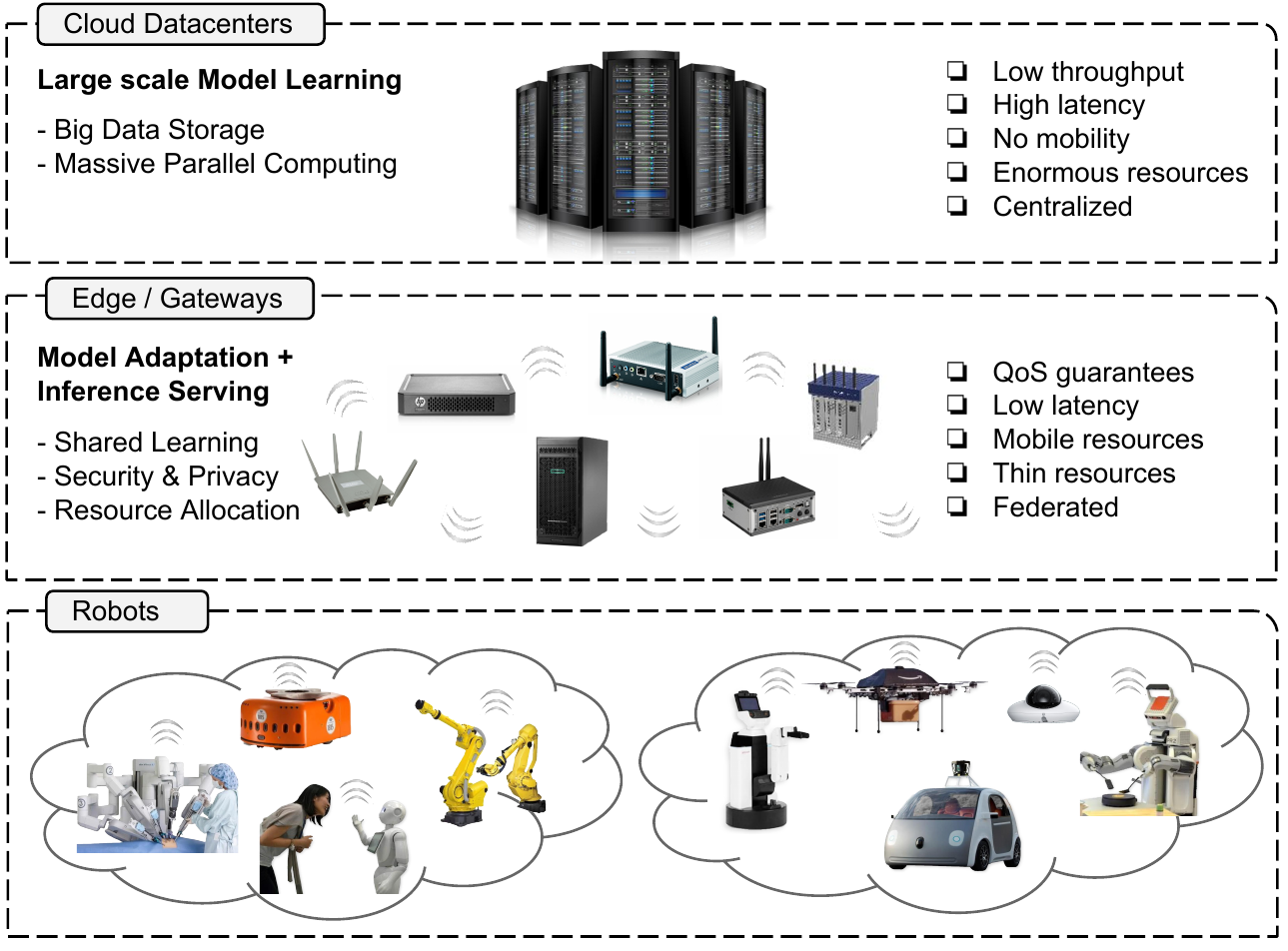}
\caption{\footnotesize A Fog Robotics approach to deep robot learning that uses resources between Cloud and Edge for training, adaptation, inference serving and updating of deep models to reduce latency and preserve privacy of the data.} \label{Fig: Fogrobotics_arch}
\end{figure}

\textbf{Fog Robotics} is \textit{``an extension of Cloud Robotics that distributes storage, compute and networking resources between the Cloud and the Edge in a federated manner''.} The term Fog Robotics (analogous to Fog Computing\footnote{The term ``Fog Computing'' was introduced by Cisco Systems in $2012$ \cite{Bonomi12}. Other closely related concepts to Fog Computing are Cloudlets \cite{Verbelen12} and Mobile Edge Computing \cite{Roman16}.}~\cite{Yi15, Dastjerdi16, Atlam18}) was first used by Gudi et al.~\cite{Gudi17}. In this paper, we apply Fog Robotics for robot learning and inference of deep neural networks such as object recognition, grasp planning, localization etc. over wireless networks. We address the system level challenges of network limits (high latency, limited bandwidth, variability of connectivity, etc.), security and privacy of data and infrastructure, along with resource allocation and model placement issues. Fog Robotics provides flexibility in addressing these challenges by: 1) sharing of data and distributed learning with the use of resources in close proximity instead of exclusively relying on Cloud resources, 2) security and privacy of data by restricting its access within a trusted infrastructure, and 3) resource allocation for load balancing between the Cloud and the Edge (see Fig.~\ref{Fig: Fogrobotics_arch} for an overview and Sec. \ref{sec: FR} for details). Shared learning reduces the burden of collecting massive training data for each robot in training deep models, while the models are personalized for each robot at the Edge of the network within a trusted infrastructure. Deploying the deep models at the Edge enables prediction serving at a low-latency of less than $100$ milliseconds.

These principles are useful to efficiently train, adapt and deploy massive deep learning models by simulation to reality transfer across a fleet of robots. Surface decluttering is a promising application of service robots in a broad variety of unstructured environments such as home, office and machine shops. Some related examples include cloud-based robot grasping~\cite{Kehoe_2013}, grasping and manipulation in home environments~\cite{Ciocarlie_2014}, robotic butler with HERB~\cite{Srinivasa09} and PR2~\cite{Bohren11}, combining grasping and pushing primitives in decluttering lego blocks with PR2~\cite{Gupta15},  and robot decluttering in unstructured home environments with low cost robots~\cite{Gupta18}. In this work, we consider decluttering scenarios where a robot learns to pick common machine shop and household objects from the floor, and place them into desired bins. We learn deep object recognition and grasp planning models from synthetic images in the Cloud, adapt the model to the real images of the robot within a trusted infrastructure at the Edge, and subsequently deploy models for low-latency serving in surface decluttering.


\subsection{Contributions}
This paper makes four contributions:

\begin{enumerate}
    \item Motivates and introduces Fog Robotics in the context of deep robot learning.
    \item Presents a deep learning based surface decluttering application and demonstrates the use of Fog Robotics compared to the alternatives of exclusive Cloud or exclusive local resources.
    \item Presents a domain invariant deep object recognition and grasping model by simulation to real transfer, and evaluates benchmarks for learning and inference with a mobile robot over a wireless network.
    \item Surface decluttering experiments with a mobile Toyota HSR robot to grasp $185$ household and machine shop objects over $213$ grasp attempts.
\end{enumerate}
\section{Fog Robotics}\label{sec: FR}

While the Cloud can be viewed as a practically infinite pool of homogeneous resources in far away data centers, the Edge of the network is characterized by a limited collection of heterogeneous resources owned by various administrative entities. Resources at the Edge come in various sizes, e.g.\ content delivery networks, light-weight micro servers, networking devices such as gateways, routers, switches and access points. Fog Robotics explores a continuum between on-board resources on a robot to far away resources in Cloud data centers. The goal is to use the available resources, both at the Edge and in the Cloud, to satisfy the service level objectives including, but not limited to, latency, bandwidth, reliability and privacy. By harnessing the resources close by and not relying exclusively on the Cloud, Fog Robotics provides opportunities such as richer communication among robots for coordination and shared learning, better control over privacy of sensitive data with the use of locally provisioned resources, and flexible allocation of resources based on variability of workload.

\textbf{Related Work: }Hong et al. proposed `Mobile Fog' to distribute IoT applications from Edge devices to the Cloud in a hierarchical manner~\cite{Hong13}. Aazam and Huh \cite{Aazam14} presented a resource allocation model for Fog Computing. Bonomi et al. made provision for resource constrained IoT devices in their Fog Computing platform \cite{Bonomi14}. In \cite{Yousefpour17}, the authors propose a framework to minimize service delays in Fog applications by load sharing. The authors in \cite{Zao14} use a multi-tier Fog and Cloud computing approach for a pervasive brain monitoring system that can reliably estimate brain states and adapt to track users' brain dynamics. Lee et al. in \cite{Lee15} and Alrawais et al. in \cite{Alrawais17} discuss the security and privacy issues and present solutions for mitigating the security threats. More details of Fog computing are in \cite{Mouradian18, Mukherjee18}. Recently, several groups have also advocated the need for Fog Robotics. Katterpur et al. profile the computation times for resource allocation in a fog network of robots~\cite{Kattepur}. Gudi et al. present a Fog Robotics approach for human robot interaction~\cite{Gudi18}. Pop et al. discuss the role of Fog computing in industrial automation via time-sensitive networking~\cite{Pop18}. For more details and updates, see~\cite{Goldberg_fog,OpenFog,SongTanwaniGoldberg19}.




As an example, a number of battery powered WiFi-enabled mobile robots for surface decluttering can use resources from a close-by fixed infrastructure, such as a relatively powerful smart home gateway, while relying on far away Cloud resources for non-critical tasks. Similar deployments of robots with a fixed infrastructure can be envisioned for industrial warehouses, self-driving cars, flying drones, socially aware cobots and so on.  Below, we review the opportunities that Fog Robotics provides for secure and distributed robot learning:

\subsection{Enabling Shared and Distributed Learning}

Fog Robotics brings computational resources closer to mobile robots that enables access to more data via different sensors on a robot or across multiple robots. Whereas Cloud Robotics assumes the Cloud as a centralized rendezvous point of all information exchange, Fog Robotics enables new communication modalities among robots by finding other optimal paths over the network. Using a Cloud-only approach is inefficient in utilizing the network bandwidth and limits the volume of data that can be shared.

Fog Robotics enables computational resources closer to the robots to perform pre-processing, filtering, deep learning, inference, and caching of data to reduce reliance on far away data centers. For example, to support household robots, models trained in the Cloud can be periodically pushed to a smart home gateway instead of directly onto individual robots; such a smart home gateway can act as a cache of local model repository, perform adaptation of a generalized model to the specific household, provide storage of data collected from the household for model adaptation, or even run a shared inference service for local robots to support robots with very limited onboard resources. We demonstrate such an inference service in the context of the surface decluttering application.

On a broader scale, resources at a municipal level allow for similar benefits at a geographical level. Such computational resources outside data centers are not merely a vision for the future; they already exist as a part of various projects such as EdgeX Foundry~\cite{edgexfoundry}, CloudLab~\cite{Ricci14}, EdgeNet~\cite{edgenet}, US Ignite~\cite{usignite}, PlanetLab~\cite{komosny2015planetlab}, PlanetLab Europe~\cite{planetlabeurope}, GENI~\cite{berman2014geni}, G-Lab~\cite{schwerdel2014future}, among others. 

\subsection{Security, Privacy, and Control Over Data}\label{sec: security_privacy}

Network connected systems significantly increase the attack surface when compared to standalone infrastructure. Deliberate disruption to wide-area communication (e.g., by targeted Denial of Service (DoS) attacks) is not uncommon~\cite{kolias2017ddos}. The control of data collected by robots and the security of data received from a remote service is a major concern. To this end, a well designed Fog Robotics application can provide a tunable middle ground of reliability, security and privacy between a `no information sharing' approach of standalone isolated deployments and a `share everything' approach of Cloud Robotics.




Such control over data, however, is non-trivial as resources at the Edge are partitioned in a number of administrative domains based on resource ownership. Heterogeneity of resources further adds to the security challenge; just keeping various software to the most up-to-date versions is cumbersome. Note that merely encrypting data may not be sufficient. As an example, simple encryption only provides data confidentiality but not data integrity---a clever adversary can make a robot operate on tampered data~\cite{song2001timing}. Moreover, addressing key-management---an integral part of cryptographic solutions---is a challenge in itself~\cite{whitten1999johnny}. Finally, managing the security of `live' data that evolves over time is more challenging than that of a static dump.

Data-centric infrastuctures such as Global Data Plane (GDP)~\cite{mor2016toward} can provide a scalable alternative to control the placement and scope of data while providing verifiable security guarantees. GDP uses cryptographically secured data containers called DataCapsules. DataCapsules are analogous to shipping containers that provide certain guarantees on data integrity and confidentiality even when they are handled by various parties during their lifetimes. The integrity and provenance of information in a DataCapsule can be verified by means of small cryptographic proofs~\cite{tamassia2003authenticated}. This allows the owners of data to restrict sensitive information in a DataCapsule to, say, a home or a warehouse. In contrast, existing Cloud storage systems (say Amazon S3) do not provide provable security guarantees and rely solely on the reputation of the Cloud provider to protect the Cloud infrastructure from adversarial infiltration. Similarly, decentralized authorization systems such as WAVE~\cite{Andersen17} can protect the secrecy of data without relying on any central trusted parties. Note that secure execution of data still remains an open challenge. A wider deployment of secure hardware such as Intel's SGX (Software Guard Extensions) technology~\cite{costan2016intel} has the potential to provide for an end-to-end security.

While it is important from an infrastructure viewpoint to maintain control over sensitive data and ensure that it does not leave the boundaries of infrastructure with known security properties, applications also need to be designed around such constraints. We demonstrate such an architecture for privacy preserving Fog Robotics scenario by using synthetic non-private data for training in the Cloud and use real-world private data only for local refinement of models.

\subsection{Flexibility of Resource Placement and Allocation}

The Cloud provides seemingly infinite resources for compute and storage, whereas resources at the Edge of the network are limited. Quality of service provisioning depends upon a number of factors such as communication latency, energy constraints, durability, size of the data, model placement over Cloud and/or Edge, computation times for learning and inference of the deep models, etc. This has motivated several models for appropriate resource allocation and service provisioning~\cite{Mukherjee18}. Chinchali et al. use a deep reinforcement learning strategy to offload robot sensing tasks over the network~\cite{Chinchali19}. Nan et al. present a fog robotic system for dynamic visual servoing with an ayschronous heartbeat signal~\cite{Tian18}. 

Flexibility in placement and usage of resources can give a better overall system design, e.g. offloading computation from the robot not only enables lower unit cost for individual robots but also makes it possible to have longer battery life. Consider, for example, a resource constrained network where GPUs are available on the Cloud and only CPUs are available at the Edge of the network. Even though a GPU provides superior computation capabilities compared to a CPU, the round-trip communication time of using a GPU in the Cloud--coupled with communication latency--vs a CPU locally is application and workload dependent. Note that the capital and operational expense for a CPU is far lower than that of a GPU. Simple application profiling may be used for resource placement in this context~\cite{Kattepur}. However, finding an appropriate balance for performance and cost is challenging when the application demands and the availability of resources keeps changing over time, making continuous re-evaluation necessary~\cite{Xu18}.

\section{Deep Learning based Surface Decluttering}
\subsection{Problem Statement} 
\begin{figure}[tbp]
\centering
\includegraphics[trim={0cm 0cm 0cm 0cm},clip,scale = 0.05]{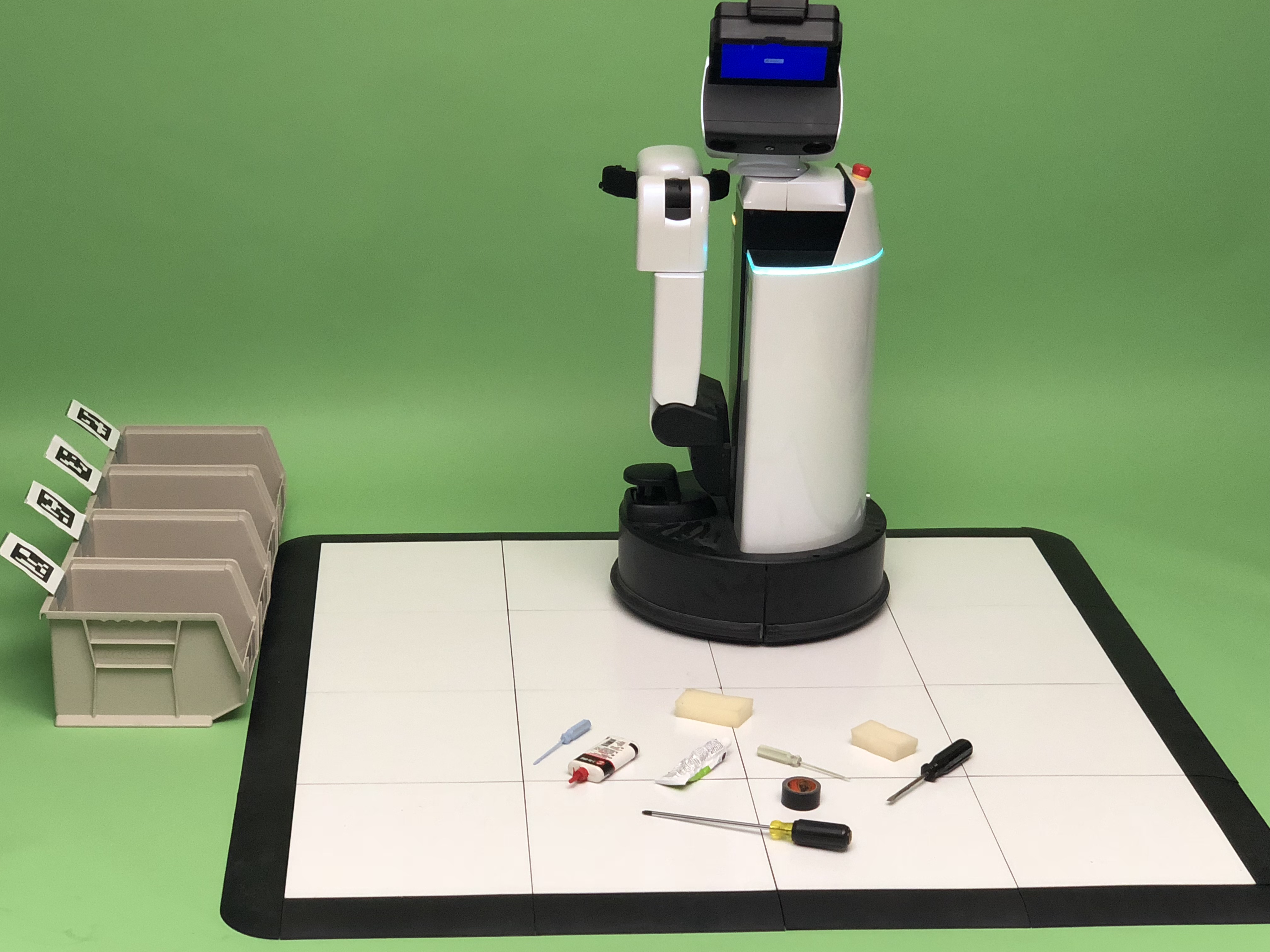}
\caption{\footnotesize Experimental setup for decluttering objects into bins with HSR.} \label{Fig: hsr_setup}
\end{figure}

We consider a mobile robot equipped with a robotic arm and a two-fingered gripper as the end-effector. The robot observes the state of the floor $\mb{\xi}_{t}$ as a RGB image $\mb{I}_{t}^{c} \in \mathbb{R}^{640 \times 480 \times 3}$ and a depth image $\mb{I}_{t}^{d} \in \mathbb{R}^{640 \times 480}$. The task of the robot is to recognize the objects $\{o_{i}\}_{i=1}^{N}$ as belonging to the object categories $o_{i} \in \{1 \ldots C\}$, and subsequently plan a grasp action $\mb{u}_t \in \mathbb{R}^{4}$ corresponding to the $3$D object position and the planar orientation of the most likely recognized object. After grasping an object, the robot places the object into appropriate bins (see Fig.~\ref{Fig: hsr_setup} for an overview).

In this paper, we learn a deep object recognition and a grasp planning model for surface decluttering with a mobile robot. The object recognition model predicts the bounding boxes of the objects from the RGB image, while the grasp planning model predicts the optimal grasp action from the depth image. We compare the grasp planning approach with a baseline that grasps orthogonal to the centroid of the principal axis of the isolated segmented objects, and uses the depth image to find the height of the object centroid. We are interested in offloading the training and deployment of these models by simulation to reality transfer with Fog Robotics. The deep models are trained with synthetic images in the Cloud, adapted at the Edge with real images of physical objects and then deployed for inference serving to the mobile robot over a wireless network.

\subsection{Simulation and Real Dataset}
We simulate the decluttering environment in a Pybullet simulator~\cite{Coumans_14}. We collect $770$ 3D object meshes from TurboSquid, KIT and ShapeNet resembling household and machine shop environments, and split them across $12$ categories: screwdriver, wrench, fruit, cup, bottle, assembly part, hammer, scissors, tape, toy, tube, and utility. Camera parameters and viewpoint in the simulator is set according to the real robot facing the floor as shown in Fig.~\ref{Fig: hsr_setup}. We randomly drop between $5-25$ objects on the floor from a varying height of $0.2-0.7$ meters, each assigned a random color from a set of $8$ predefined colors. The objects are allowed to settle down before taking the RGB and the depth image and recording the object labels. We generated $20 K$ synthetic images of cluttered objects on the floor following this process. 

\begin{figure}[tbp]
\centering
\includegraphics[trim={0cm 6.5cm 6.0cm 0.0cm},clip,scale = 0.107]{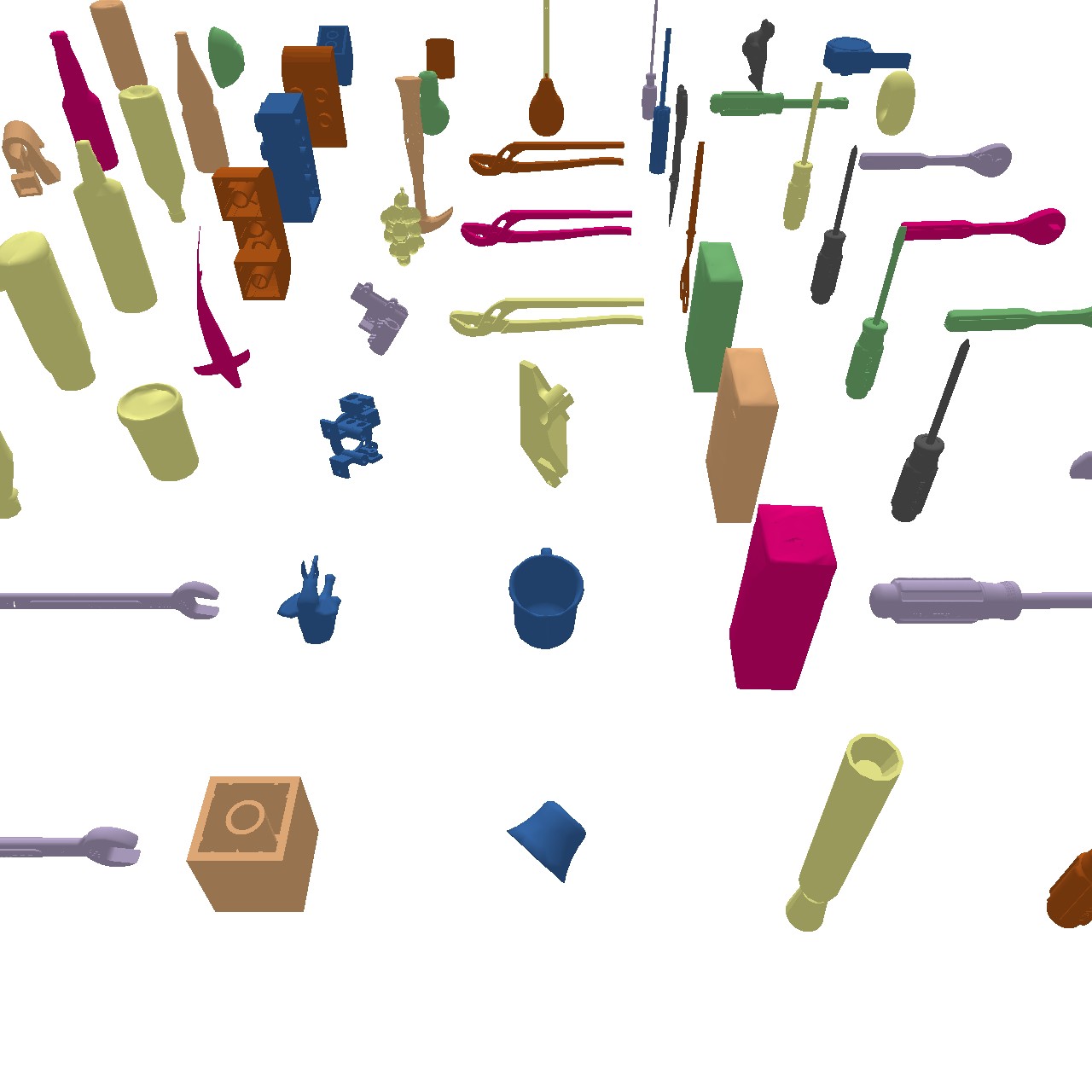} \rulesep
\includegraphics[trim={0cm 0.1cm 0.1cm 0.25cm},clip,scale = 0.039]{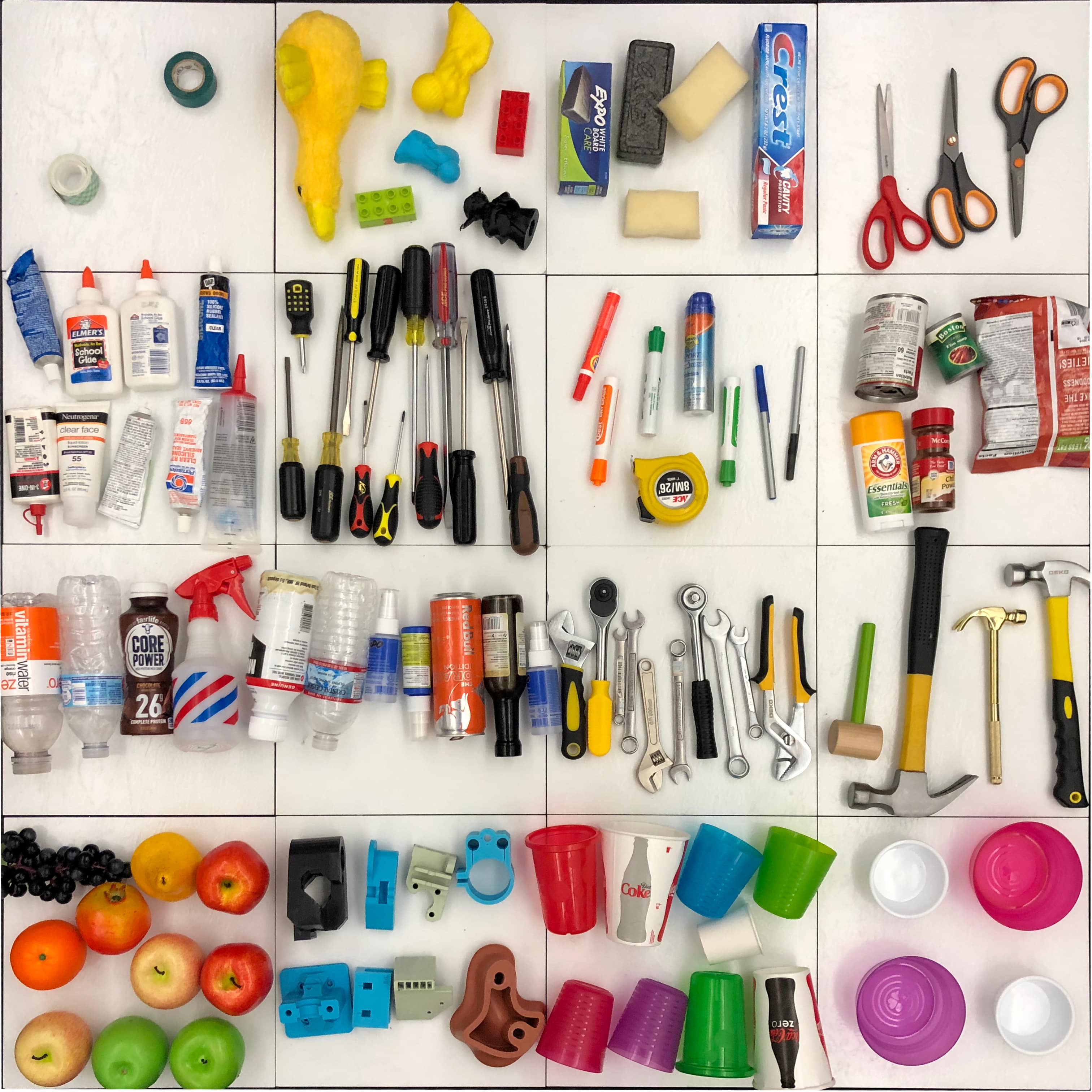} 
\caption{\footnotesize Simulation object models on \textit{(left)} and physical objects on \textit{(right)} for decluttering.} \label{Fig: obj_dataset}
\end{figure}

The physical dataset includes $102$ commonly used household and machine shop objects split across $12$ class categories as above (see Fig. \ref{Fig: obj_dataset}). We randomly load $5-25$ objects in a smaller bin without replacement and drop them on $1.2$ sq. meter white tiled floor from different positions. We collected $212$ RGB and depth camera images with an average number of $15.4$ objects per image, and hand label the bounding box and the image categories.


\subsection{Transfer Learning from Simulation to Reality} \label{sec: obj_rec}
We train the deep object recognition model on simulated data and adapt the learned model on real data such that the feature representations of the model are invariant across the simulator and the real images~\cite{Ganin16}. The learning problem considers the synthetic images as belonging to a non-private simulated domain $D_{S}$, and real images belonging to a private real domain $D_{R}$ that is not to be shared with other networks. The simulated and real domain consists of tuples of the form $D_S = \{\mb{\xi}_t^{(s)}, \mb{u}_t^{(s)}, \mb{y}_t^{(s)}\}_{t=1}^{T_S}$ and $D_R = \{\mb{\xi}_t^{(r)}, \mb{u}_t^{(r)}, \mb{y}_t^{(r)}\}_{t=1}^{T_R}$, where $\mb{y}_t^{(s)}$ and $\mb{y}_t^{(r)}$ correspond to a sequence of bounding boxes of object categories as ground-truth labels for a simulated image and a real image respectively, and $T_S \gg T_R$. The real images and the synthetic images may correspond to different but related randomized environments such as a machine shop and a household environment. For example, we randomize the colors of the 3D object models in the simulated domain, but real world objects have a fixed texture. 

We use the MobileNet-Single Shot MultiBox Detector (SSD) \cite{Liu15,Lin17} algorithm with focal loss and feature pyramids as the base model for object recognition (other well-known models include YOLO, Faster R-CNN; see \cite{Huang16} for an overview). We modify the base model such that the output of feature representation layer of the model is invariant to the domain of the image, i.e., $\mb{\xi}_t \sim D_S \approx \mb{\xi}_t \sim D_R$, while minimizing the classification loss $\mathcal{L}_{y_c}$ and the localization loss $\mathcal{L}_{y_l}$ of the model. We add an adversarial discriminator at the output of the feature representation layer that predicts the domain of the image as synthetic or real $\mb{\xi}_t \in \{D_S,D_R\}$. The overall model parameters are optimized such that the the object classification loss $\mathcal{L}_{y_c}$ and the localization loss $\mathcal{L}_{y_l}$ is minimized, while the domain classifier loss $\mathcal{L}_{d}$ is maximally confused in predicting the domain of the image~\cite{Goodfellow14,Ganin16,Tzeng17}. The trade-off between the loss functions governs the invariance of the model to the domain of the image and the output accuracy of the model. 

We denote the augmented model as the \textit{domain invariant object recognition} model (DIOR). The domain classification architecture is empirically selected to give better performance with $3$ fully connected layers of $1024, 200$ and $100$ neurons after flattening the output of logits layer. We implement two variations of DIOR: 1) \textbf{DIOR\_dann} that shares parameters of the feature representation for both the sim and the real domain~\cite{Ganin16}, and 2) \textbf{DIOR\_adda} that has separate feature representation layers for the sim and the real domain to allow different feature maps. The parameters of the sim network are pretrained and fixed during the adaptation of the real images in this variant~\cite{Tzeng17}. 

The cropped depth image from the output bounding box of the object recognition model is fed as input to the Dex-Net grasp planning model adapted from~\cite{mahler2017dex}. The model is retrained on synthetic depth images as seen from the tilted head camera of the robot in simulation. Depth images are readily invariant to the simulator and the real environment. The grasp planning model samples antipodal grasps on the cropped depth image of the object and outputs the top ranked grasp for the robot to pick and place the object into its corresponding bin. 

\subsection{Networked System with Execution Environments}

\begin{figure}[tbp]
\centering
\includegraphics[trim={0.0cm 0.0cm 0cm 0cm},clip,scale = 0.50]{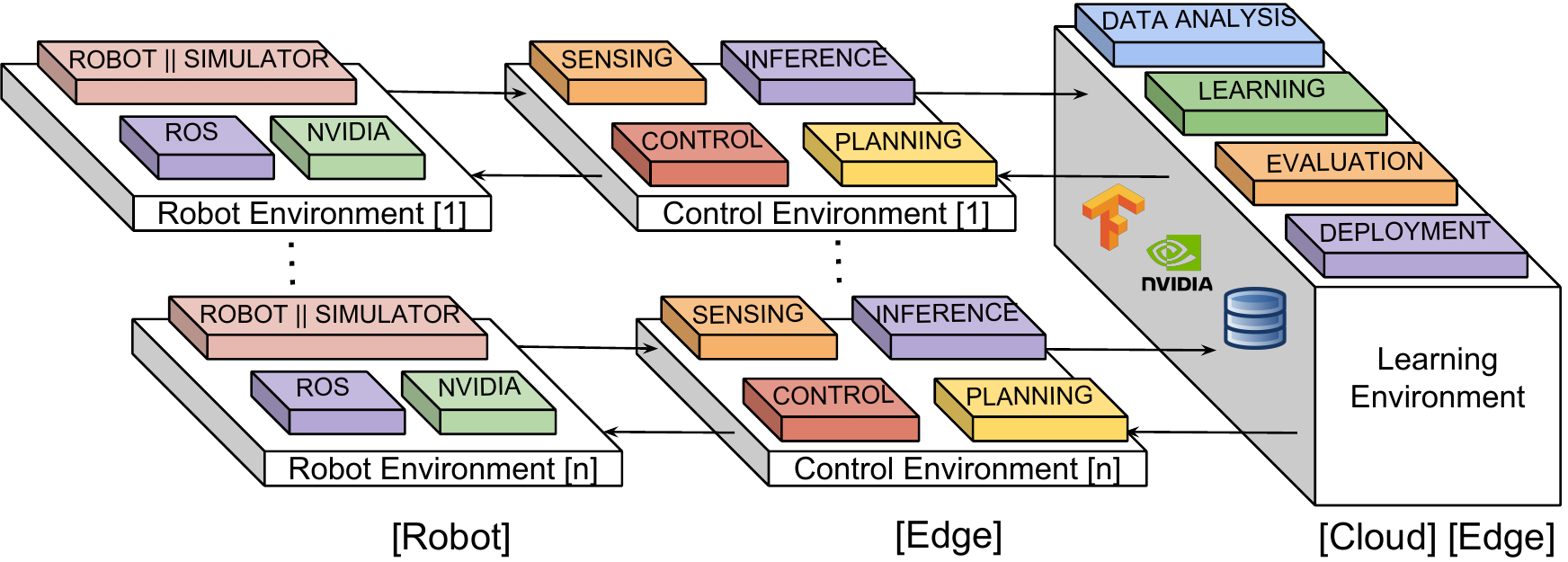}
\caption{\footnotesize Software components running in network-connected execution environments packaged and distributed via Docker images: \textit{(left)} robot environment, \textit{(centre)} control environment, \textit{(right)} learning environment.} \label{Fig: docker_arch}
\end{figure}

The overall networked system consists of three modular execution environments (see Fig.~\ref{Fig: docker_arch}): 1) the robot environment or its digital twin~\cite{Kirill17} in the simulator that sends images of the environment and receives actuation commands to drive the robot; 2) the control environment responsible for sensing the images, \textit{inferring} the objects and grasp poses from the images using the trained object recognition and grasp planning model, planning the motion of the robot for executing the grasps, and sending the actuation commands to drive the robot; and 3) the learning environment that receives images and labels from the robot or the simulator and splits the data for training and evaluation of the deep models. At the end of the training process, the best performing model on the evaluation set is deployed as an \textit{inference graph} for secured and low-latency prediction serving at the Edge in the \textit{robot-learning-as-a-service} platform. The platform defines a service for robots to easily access various deep learning models remotely over a gRPC server. Note that the robot environment, the control environment and the learning environment are virtual, and their appropriate placement depends on the available storage and compute resources in the network. The software components running in network-connected execution environments are packaged and distributed via Docker images~\cite{Merkel14}. 


We run an instance of the learning environment to train the deep object recognition model on the Cloud with the non-private synthetic data only, while another instance runs at the Edge of the network that adapts the trained network on real data to extract invariant feature representations from the private (real) and the non-private (synthetic) data.

\section{Experiments, Results and Discussion}
We now present comparative experiments of deep learning and inference for surface decluttering using: 1) Cloud resources only, 2) Edge resources only, and 3) Fog using resources on both Cloud and Edge. The Edge infrastructure includes a workstation (6-core, 2-thread Intel CPUs, 1.1 TB Hard Disk, with a Titan XP GPU) located in UC Berkeley for Edge computing and storage. We use the Amazon EC$2$ \texttt{p$2.8$xlarge} instance with $8$ Tesla K$80$ GPUs for Cloud compute and use Amazon S3 buckets for Cloud storage. We launch the EC2 instance in two regions: 1) \textbf{EC2 (West)} in Oregon (us-west-2), and 2) \textbf{EC2 (East)} in Northern Virginia (us-east-1). 

\subsection{Sim-to-Real Domain Adaptation over the Network}

We divide both the simulated and the real datasets into $60 \%$ training and $40 \%$ evaluation sets: $\{\mathrm{sim\_train, real\_train, sim\_eval, real\_eval}\}$, and estimate the model parameters described in Sec. \ref{sec: obj_rec} under different networks on $\mathrm{real\_eval}$: 1) training in the \textbf{Cloud} with only large scale non-private synthetic images $\{\mathrm{sim\_train}\}$, 2) training at the \textbf{Edge} with only limited number of private real images $\{\mathrm{real\_train}\}$, and 3) training in the \textbf{Fog} with both synthetic and real images on the Edge, using a pretrained model on large scale synthetic data in Cloud, under $3$ baselines: a) \textbf{Sim+Real: }training on combined simulation and real data with no domain classifier, b) \textbf{DIOR\_dann: }training DIOR with shared parameters for sim and real feature representation, c) \textbf{DIOR\_adda: } training DIOR with separate parameters for sim and real feature representations.

Results are summarized in Table \ref{tab_Cloud_Fog_dl}. We observe that the models give comparable or better mean average precision (mAP)~\cite{Lin14} and classification accuracy on the real images in the Fog in comparison to the models trained exclusively on Cloud or Edge. Naive transfer of model trained on synthetic data does not perform well on the real data with an accuracy of $24.16\%$. Combining sim and real data naively is also suboptimal. The domain invariant object recognition with a few labeled real images provides a trade-off between acquiring a generalized representation versus an accurate adaptation to the real images. The \textbf{DIOR\_adda} drastically improves the performance on real domain by partially aligning the feature representation with the sim domain. The \textbf{DIOR\_dann} model with shared parameters gives good performance in both domains, which can further be used to update the simulator model in the Cloud~\cite{Chebotar18}. We report the remainder of the results with \textbf{DIOR\_dann}. Training time of each model is over $13$ hours on both the Cloud and the Edge(GPU) instances suggesting that the model placement issues are less critical for training. 

\begin{figure}[tbp]
\centering
\includegraphics[trim={2.6cm 3.4cm 2.2cm 3.9cm},clip,width = 6.25cm, height=3.9cm]{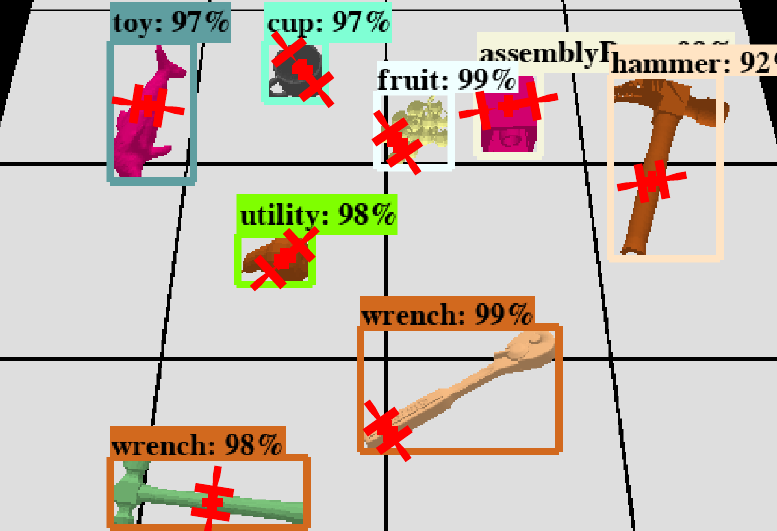}

\vspace{1mm}

\includegraphics[trim={2.9cm 3.0cm 2cm 4.6cm},clip,scale = 0.8]{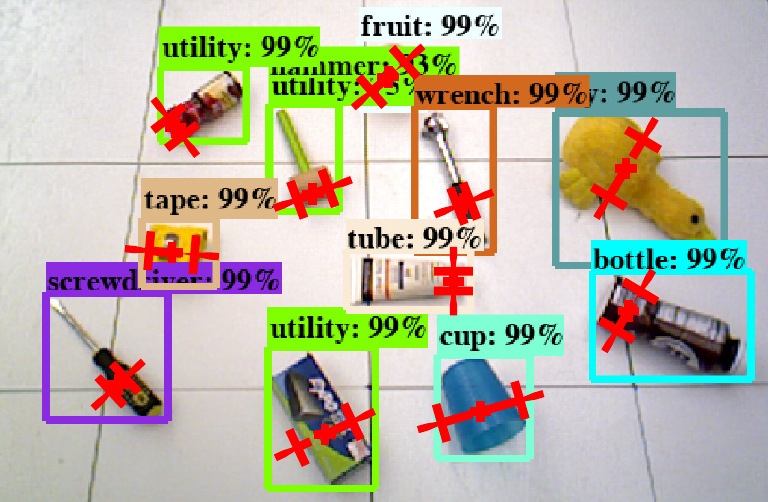}

\caption{\footnotesize Object recognition and grasp planning model output on a simulated image on \textit{(top)} and real image on \textit{(bottom)} as seen from the robot head camera.} \label{Fig: results_obj_rec}
\end{figure}

The cropped depth image of the closest object to the robot is fed to the grasp planning model to compute the grasp poses for robot surface decluttering (see Fig. \ref{Fig: results_obj_rec} for qualitative results of the model on both synthetic and real data). 

\begin{table}[tb]
\caption{\footnotesize Comparative experiments for learning deep object recognition for simulation to reality transfer over Cloud, Edge and Fog. Metrics include mean Average Precision (mAP) on real images, classification accuracy on synthetic test images $\mathrm{sim\_eval}$, real test images $\mathrm{real\_eval}$ and both synthetic and real test images $\mathrm{mix\_eval}$. Domain invariant object recognition with shared feature representation network parameters $\mathrm{DIOR\_dann}$ model gives better performance in both simulation and real domain using Fog Robotics.} \normalsize \centering \label{tab_Cloud_Fog_dl}
\begin{tabular}{|c||c|c|c|c|}
\hline
Training Set & mAP & $\mathrm{sim\_eval}$ & $\mathrm{real\_eval}$ & $\mathrm{mix\_eval}$\\ \hline \hline
\multicolumn{5}{|c|}{\textbf{Cloud}}\\\hline
\textbf{Sim} & $0.13$ & $97.97$ & $24.16$  & $55.5$\\ \hline \hline
\multicolumn{5}{|c|}{\textbf{Edge}}\\\hline
\textbf{Real} & $0.62$ & $24.64$ & $88.1$& $64.92$ \\ \hline \hline
\multicolumn{5}{|c|}{\textbf{Fog}}\\\hline
\textbf{Sim $+$ Real} & $0.33$& $90.40$ & $54.12$ & $69.97$ \\ \hline
\textbf{DIOR\_dann} & $0.61$& $96.92$& $86.33$ & $\mb{95.21}$ \\ \hline
\textbf{DIOR\_adda} & $0.61$& $30.87$& $\mb{90.64}$ & $67.82$ \\ \hline
\end{tabular}
\end{table}


\subsection{Communication vs Computation Cost for Inference}


We deployed the trained models in the robot-learning-as-a-service platform that receives images from the robot as a client, performs inference on a server, and sends back the result to the robot. We measure the round-trip time $t^{\mathrm{(rtt)}}$, i.e., time required for communication to/from the server and the inference time $t^{\mathrm{(inf)}}$. We experiment with four hosts for the inference service in the order of decreasing distance to the robot: EC2 Cloud (West), EC2 Cloud (East), Edge with CPU support only, and Edge with GPU support. 

Results in Table~\ref{tab_inference} show that the communication and not the computation time is the major component in overall cost. Deploying the inference service  on the Edge significantly reduces the round-trip inference time and the timing variability in comparison to hosting the service on Cloud, with nearly $4\times$ difference between EC2 Cloud host (East) and Edge host with GPU. 



\subsection{Surface Decluttering with the Toyota HSR}

We test the performance of the trained models on the mobile Toyota HSR robot for surface decluttering. We load $5-25$ objects in a smaller bin from a smaller set of $65$ physical objects and drop them on the floor in front of the robot (see Fig. \ref{Fig: hsr_setup}). The overall accuracy of the domain invariant object recognition and the grasping model on the robot is $90.14\%$ and $86.85\%$, respectively, for a total of decluttering $185$ objects across $213$ grasp attempts. In comparison, grasping orthogonal to the principal axis of the segmented objected resulted in a grasping accuracy of $76.19\%$ only. We found that the grasping results improved substantially by retraining the model with respect to the tilted camera viewpoint of the robot in comparison to the results reported in~\cite{Gupta18}. Note that we remove the pathological objects such as heavy hammers, and objects with very low ground clearance such as wrenches and scissors that the robot is not able to grasp. We observe that the robot performs well in grasping compliant objects and objects with well-defined geometry such as cylinders, screwdrivers, tape, cups, bottles, utilities, and assembly parts (see \todo{\url{https://sites.google.com/view/fogrobotics}} for video, results and supplementary details).

\begin{table}[tb]
\caption{\footnotesize Computation time for inference $t^{\mathrm{(inf)}}$ vs round trip communication time $t^{\mathrm{(rtt)}}$ (in milliseconds) for inference over Edge (with and without GPU) and Cloud with EC2 (West) EC2 (East) instances. Results are averaged across $200$ real images. Communication time dominates the computation time and increases as the distance to the server increases.} \normalsize \centering \label{tab_inference}
\begin{tabular}{|c||c|c|}
\hline
 Location & $t^{\mathrm{(inf)}}$ &
 $t^{\mathrm{(rtt)}}$ \\\hline \hline
 \multicolumn{3}{|c|}{\textbf{Object Recognition}}\\\hline
 \textbf{EC2(East)} & $31.93 \pm 1.53$ & $437.63 \pm 100.02$ \\\hline 
 \textbf{EC2(West)} & $\mb{31.12 \pm 1.28}$ & $181.61 \pm 22.71$ \\\hline
 \textbf{Edge(CPU)} & $52.34 \pm 4.18$ & $149.32 \pm 21.04$ \\\hline
 \textbf{Edge(GPU)} & $33.27 \pm 3.09$ & $\mb{119.40 \pm 12.06}$ \\\hline \hline
 \multicolumn{3}{|c|}{\textbf{Grasp Planning}}\\\hline
 \textbf{EC2(East)} & $1906.59 \pm 224.19$ & $ 4418.34 \pm 1040.59$ \\\hline
 \textbf{EC2(West)} & $1880.28 \pm 207.46 $ & $ 2197.76 \pm 199.44$ \\\hline
 \textbf{Edge(CPU)} & $ 3590.71 \pm 327.57$ & $ 3710.74 \pm 214.08$ \\\hline
 \textbf{Edge(GPU)} & $\mb{1753.65 \pm 201.38}$ & $\mb{1873.16 \pm 211.57}$ \\\hline
\end{tabular}
\end{table}

\section{Conclusions and Future Directions}
In this paper, we have introduced a Fog Robotics approach for secure and distributed deep robot learning. Secured compute and storage at the Edge of the network opens up a broad range of possibilities to meet lower-latency requirements, while providing better control over data. Standardizing robot communication with available Edge resources, nonetheless, is challenging for a wider adoption of Fog Robotics. We have presented a surface decluttering application, where non-private (public) synthetic images are used for training of deep models on the Cloud, and real images are used for adapting the learned representations to the real world in a domain invariant manner. Deploying the models on the Edge significantly reduces the round-trip communication time for inference with a mobile robot in the decluttering application. In future work, we plan to deploy various deep models for segmentation, hierarchical task planning etc, for low-latency and secure prediction in a multi-agent distributed environment with a set of robots.

\section*{ACKNOWLEDGMENT}
\footnotesize
This research was performed at the AUTOLAB at UC Berkeley in affiliation with the Berkeley AI Research (BAIR) Lab, Berkeley Deep Drive (BDD), the Swarm Lab, the Real-Time Intelligent Secure Execution (RISE) Lab, the CITRIS ``People and Robots'' (CPAR) Initiative, and by the Scalable Collaborative Human-Robot Learning (SCHooL) Project, NSF National Robotics Initiative Award 1734633, and the NSF ECDI Secure Fog Robotics Project Award 1838833. The work was supported in part by donations from Siemens, Google, Amazon Robotics, Toyota Research Institute, Autodesk, ABB, Knapp, Loccioni, Honda, Intel, Comcast, Cisco, Hewlett-Packard and by equipment grants from PhotoNeo, NVidia, and Intuitive Surgical. The authors would like to thank Flavio Bonomi, Moustafa AbdelBaky, Raghav Anand, Richard Liaw, Daniel Seita, Sanjay Krishnan and Michael Laskey for their helpful feedback and suggestions.

\small
\bibliographystyle{IEEEtran}
\balance
\bibliography{0-bibliography-short}

\end{document}